\UseRawInputEncoding
\documentclass[letterpaper, 10 pt, conference]{ieeeconf}  % Comment this line out if you need a4paper

\usepackage{graphicx}
\usepackage{hyperref}
\usepackage{siunitx}
\IEEEoverridecommandlockouts                              % This command is only needed if 
\title{\LARGE \bf
An Experimental Study of Wind Resistance and Power Consumption in MAVs with a Low-Speed Multi-Fan Wind System
}

\author{Diana A. Olejnik$^{1}$, Sunyi Wang$^{1}$, Julien Dupeyroux$^{1}$, Stein Stroobants$^{1}$,  Mat\v{e}j Kar\'{a}sek\ $^{1}$, Christophe De Wagter$^{1}$ \\ and Guido de Croon$^{1}$% <-this % stops a space
\thanks{$^{1}$All authors are with the MAVLab, Department of Control and Operations, TU Delft, the Netherlands. Contact author:
        {\tt\small diana.olejnik@gmail.com}}%
}

\begin{document}

\maketitle
\thispagestyle{empty}
\pagestyle{empty}

%%%%%%%%%%%%%%%%%%%%%%%%%%%%%%%%%%%%%%%%%%%%%%%%%%%%%%%%%%%%%%%%%%%%%%%%%%%%%%%%
\begin{abstract}

This paper discusses a low-cost, open-source and open-hardware design and performance evaluation of a low-speed, multi-fan wind system dedicated to micro air vehicle (MAV) testing. In addition, a set of experiments with a flapping wing MAV and rotorcraft is presented, demonstrating the capabilities of the system and the properties of these different types of drones in response to various types of wind. We performed two sets of experiments where a MAV is flying into the wake of the fan system, gathering data about states, battery voltage and current. Firstly, we focus on steady wind conditions with wind speeds ranging from 0.5 m/s to 3.4 m/s. During the second set of experiments, we introduce wind gusts, by periodically modulating the wind speed from 1.3 m/s to 3.4 m/s with wind gust oscillations of 0.5 Hz, 0.25 Hz and 0.125 Hz. The "Flapper" flapping wing MAV requires much larger pitch angles to counter wind than the "CrazyFlie" quadrotor. This is due to the Flapper's larger wing surface. In forward flight, its wings do provide extra lift, considerably reducing the power consumption. In contrast, the CrazyFlie's power consumption stays more constant for different wind speeds. The experiments with the varying wind show a quicker gust response by the CrazyFlie compared with the Flapper drone, but both their responses could be further improved. We expect that the proposed wind gust system will provide a useful tool to the community to achieve such improvements.
\end{abstract}

%%%%%%%%%%%%%%%%%%%%%%%%%%%%%%%%%%%%%%%%%%%%%%%%%%%%%%%%%%%%%%%%%%%%%%%%%%%%%%%%
\section{Introduction}

During the last years, we observe a growing demand for drones from the commercial and government sectors. Technological advances in sensory and processing capabilities of micro air vehicles (MAVs) bring them closer to their bigger size counterparts. For inspecting narrow or cluttered environments in a safe manner, small and agile MAVs are the best choice. However, due to their small size and low weight, MAVs are heavily affected by even modest wind gusts. Hence, there is an emerging need for algorithms and design solutions that will ensure efficient and stable flight of MAVs, even in windy conditions. Modeling aerodynamics is still an active area of research, e.g., for flapping wing drones with complex fluid-to-structure interactions \cite{c15,c16} or for rotorcraft in confined spaces \cite{c17,c18}. Hence, in order to test and evaluate wind resistance simulation studies are insufficient. An experimental approach is needed, preferably with a controlled, indoor wind generation setup.
Existing experimental approaches are typically taking advantage of large, expensive wind tunnel facilities. These facilities generally produce steady wind with a laminar flow that  does not represent well real outdoor scenarios. The nature of wind is unstable and gusty. Hence there is a need for wind systems that can generate variable wind conditions. Gust generation by utilising multiple oscillating vanes is a common experimental method for simulating atmospheric gusts \cite{c1}. With the aim of performing gust response experiments on aircraft wings \cite{c2} developed a low subsonic wind tunnel gust generator based on two rectangular gust vanes with a symmetric airfoil oscillating in pitch. \cite{c3} presented a similar solution but with only one gust vane. Another approach assumes an active grid installed on the wind tunnel nozzle \cite{c4} or the use of blowing air jets mounted on two fixed profiles as in \cite{c5}. In contrast, passive solutions for turbulence generation assume the use of roughness blocks, spires and grids mounted in the wind tunnel. With those, it is hard to simulate low-frequency fluctuations \cite{6}.

\begin{figure}[t!]
      \centering
      \includegraphics[width = \columnwidth]{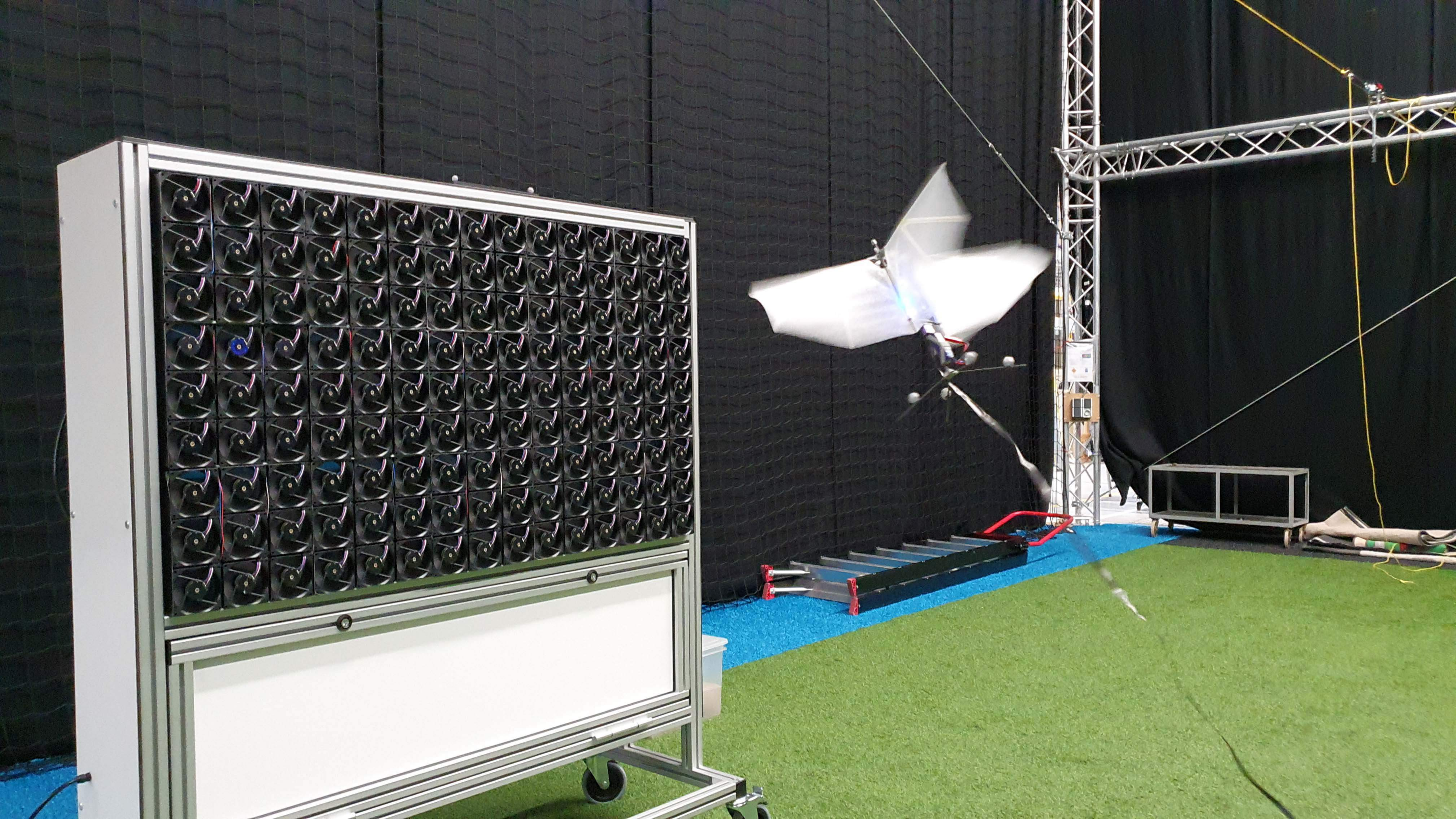}
      \caption{Open-source, open-hardware multi-fan wind system dedicated to MAVs testing.}
      \label{intro}
  \end{figure}
  
The next category of wind gust generators is that of multiple-fan-based systems. \cite{c7} demonstrates the performance of 99 fans mounted in an open-circuit wind tunnel. \cite{c8} proposes multiple fans mounted on swivel plates to easily control wind speed and direction. \cite{c9} presents a multi-fan wind tunnel concept integrated into a low pressure environmental chamber at NASA’s Jet Propulsion Laboratory to simulate the external free stream flow of forward-flight scenarios in a low-density environment. Finally, also a commercial product was developed by the company "Windshape", a multiple-fan system that promises real wind and weather imitation \cite{c10}.

The majority of solutions presented here require a sophisticated facility to generate wind gusts or/and inevitably entail exceedingly high investment costs. Multiple-fan systems have an advantage due to the possibility to independently control each fan and are more effective in generating a large turbulence scale. Thus, they can mimic real-world scenarios more closely. Moreover, in the case of bio-inspired flapping-wing MAVs, such a system will allow for a higher similarity between animal and robot experiments.

Hence, we present here, a low-speed multi-fan wind system dedicated to MAVs testing (See Figure~\ref{intro}). Our main contribution here is describing the low-cost, mobile and open-source system, its design and performance evaluation. In this way, we hope to make it accessible for a bigger group of researchers. In addition, we performed experiments with a flapping wing MAV and rotorcraft to demonstrate the usefulness and capabilities of the system. We assumed two sets of experiments, where a drone is flying into the wake of the fan system, gathering data about states, battery voltage and current. Firstly, we focus on steady wind conditions. We set a wind speed and record around 50 seconds of free flight. We repeat the experiment six times for wind speeds ranging from 0.5 m/s to 3.4 m/s. During the second set of experiments, we introduce a wind gust, by periodically modulating wind speed from 1.3 m/s to 3.4 m/s in a loop. The tested frequencies of wind gust oscillations are 0.5 Hz, 0.25 Hz and 0.125 Hz. 

\section{The wind system design and development}

\subsection{Working principle and design}

\begin{figure}[h!]
      \centering
      \includegraphics[width=0.9\columnwidth]{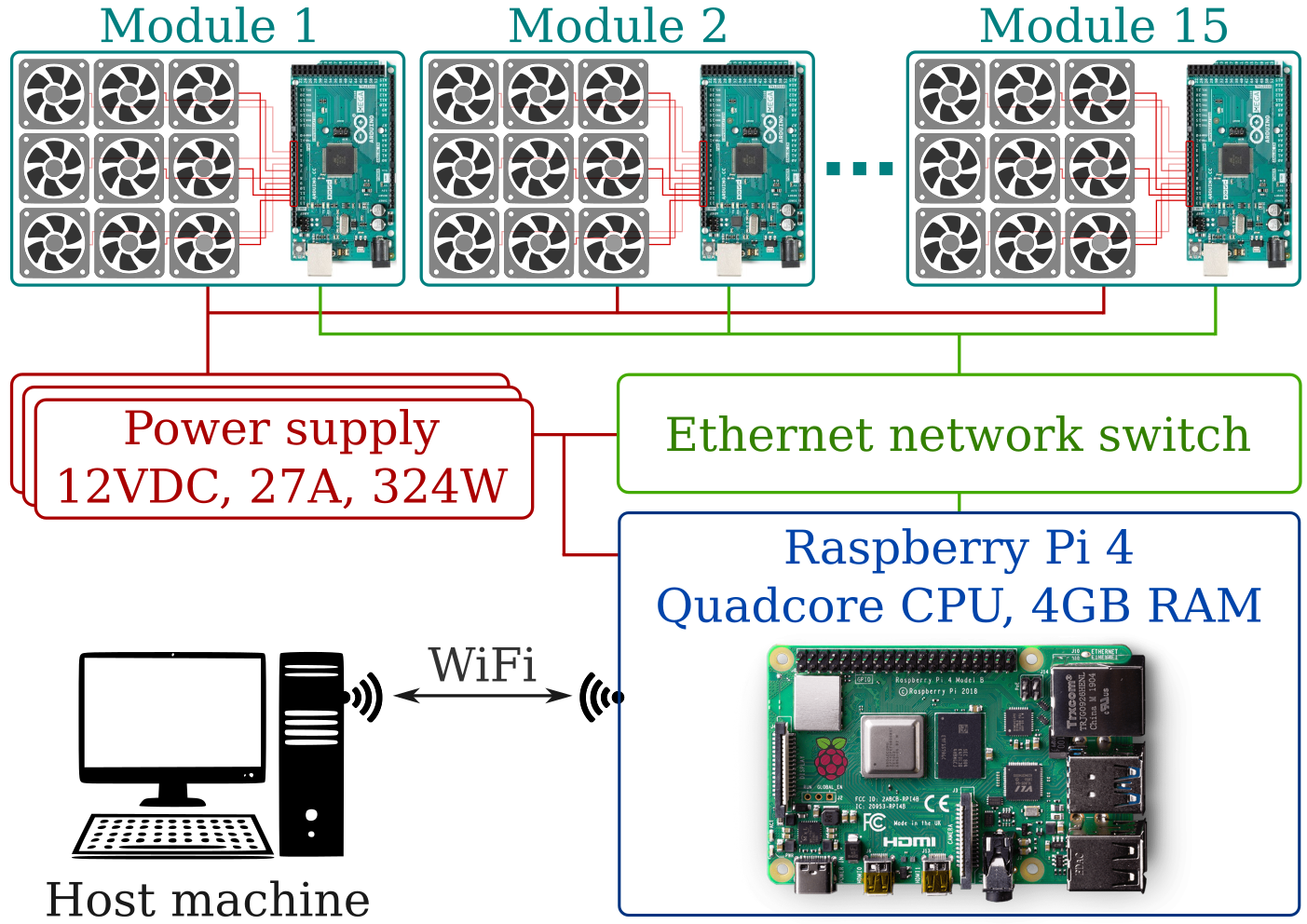}
      \caption{General overview of the hardware architecture of our wind system. Each wind module consists of 9 fans controlled by an Arduino Mega microcontroller via PWM signals. All modules are controlled by a script running on the Raspberry Pi 4 board via a UDP protocol established over an Ethernet network. The user can define and change the parameters of the wind generation from a host machine connected to the Raspberry Pi via a standard WiFi protocol.}
      \label{fig:hardware}
      \end{figure}

For the purpose of drone testing in real-life scenarios, a low-speed multi-fan wind system was designed and built. The solution presented in this study consists of an array of 135 axial fans that occupy a space of 1.2 x 0.75 meters, for a total wind surface of 0.9m$^2$. The 4-wire fans offer a pulse width modulation (PWM) based control and access to a tachometer reading. The system is divided into 15 modules, where each module comprises 9 fans along with an independent control unit, a low-power CMOS 8-bit microcontroller based on the AVR Arduino Mega architecture. All modules are equipped with an Ethernet port in order to allow connection to a network switch. Next in the line is Raspberry Pi 4B (RPi), which collects and sends data periodically through the UDP protocol. A simple Python script allows to set a certain PWM signal, read RPM of each fan and set preprogrammed functions for oscillating wind airstreams at various frequencies and RPMs ranging from 0-3600. The hardware architecture is shown in Fig.~\ref{fig:hardware}. The electrical components like power supplies, fuses, network switch and RPi are mounted on a DIN rail. The entire system is designed to be powered from a non-industrial power outlet (12V/500W at full-load operation). The enclosure of the fan was designed in `Autodesk Professional' and constructed using aluminium profiles, metal sheets, HPL (High Pressure Laminate) plates. The system is modular and allows to simply access, remove or replace sections of 9 fans mounted on a box-like frame, driven by telescopic guides. The enclosure placed on a set of sturdy wheels makes it a mobile unit that is easily transportable. The wind system is an open-source project that can be accessed through the following GitHub repository: \url{https://github.com/tudelft/wind\_system}.

\subsection{Performance}

\subsubsection{Quantitative flow visualization}

\begin{figure}[h]
      \centering
      \includegraphics[scale = 0.055]{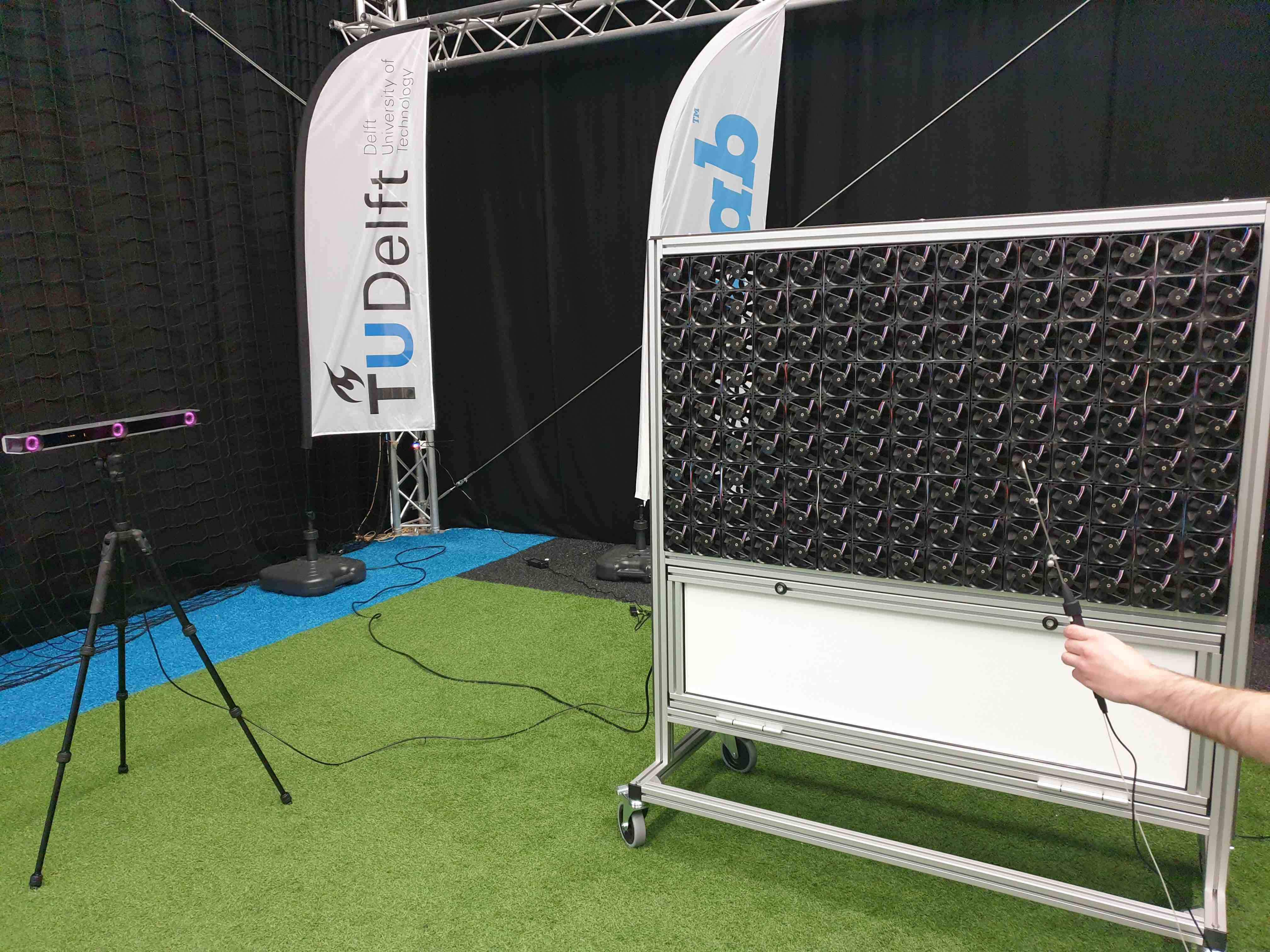}
      \caption{Experimental setup of the wind speed mapping using the Probe Capture (ProCap) tool with the infrared cameras. The 7-hole pressure probe is equipped with three reflective passive markers to allow position tracking.}
      \label{procap}
   \end{figure}

For a preliminary assessment of the performance of the wind system, we have used the Probe Capture (ProCap) tool, a device for instant flow visualization and measurement. We have mapped the airflow of the fan system by manually scanning the region in front of the fan array by the measuring device. The 7-hole pressure probe equipped with three passive markers tracked by a motion tracking system, in this case, an OptiTrack system, allows us to acquire an instantaneous position of the probe and to process the measured flow field in real-time. Figure~\ref{procap} shows the experimental setup. 
 
\begin{figure}[h]
      \centering
      \includegraphics[width = 0.9\columnwidth]{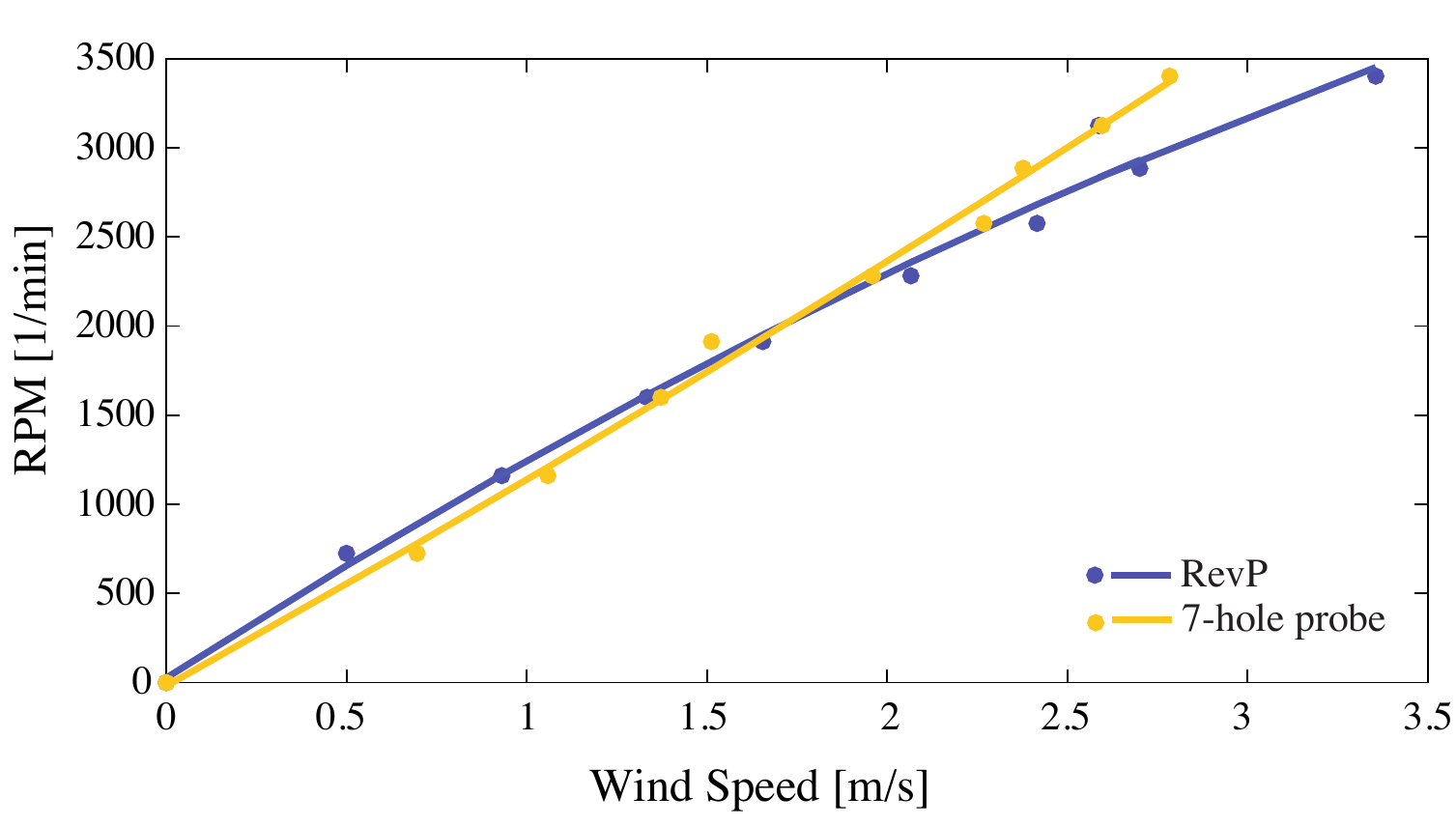}
      \caption{Time-averaged wind speed mapping of the fan system at various RPMs, 1 meter from the fan array. The measurement taken with the 7-hole probe (the ProCap tool) is marked in green and the RevP sensor-based sensing grid is marked in blue. }
      \label{fig:RPM}
      \end{figure}

The flow was measured at three streamwise positions from the fan array: 0.5, 1 and 1.5 meters from the fan array. Post-processing and comparison of the interpolated data were performed in Paraview, an open-source software for multi-platform data analysis and visualization applications. Figure \ref{fig:RPM} presents the obtained mapping of the RPM in a function of the average wind speed measured at a distance of 1 meter.

\subsubsection{Flow uniformity investigation}
\begin{figure}[h!]
      \centering
      \includegraphics[scale = 0.25]{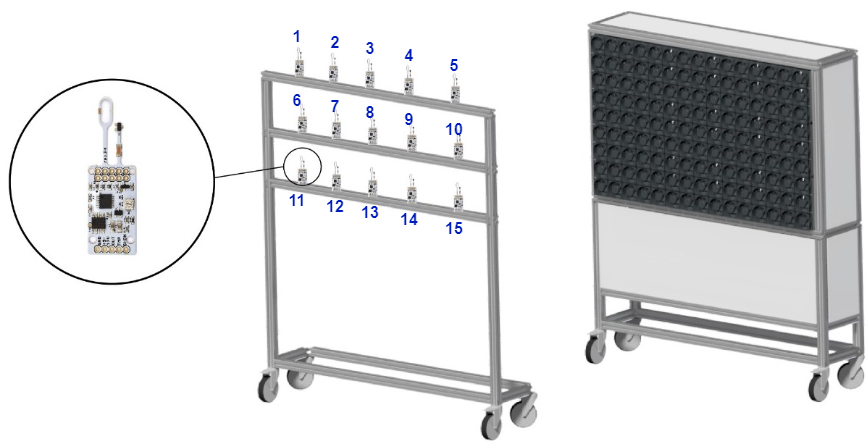}
      \caption{Test set-up for flow field mapping of the fan system at various RPMs, with 15 RevP sensors forming a wind sensing grid.}
      \label{fig:RevP_grid_setup}
      \end{figure}

To further examine the flow field uniformity and turbulence level at the 1-meter optimal downstream distance for MAV operation, 15 RevP airflow sensors from Modern Device form a sensing grid that is placed in front of the fan system as illustrated in Figure \ref{fig:RevP_grid_setup}. These highly sensitive thermistor-based airflow sensors are calibrated using a static method in the wind tunnel at the TUDelft aerodynamics facility, at a room temperature of approximately 21$^{\circ}$C, which is consistent with the temperature during the flow tests of the fan system. The convenient setup and flexible installation of the low-cost sensor grid make it a robust tool to accompany the fan system when flow quality needs to be examined. Inspired by hot-wire anemometer calibration, an air velocity calibrator from TSI (Model 1127) is used to improve the RevP calibration accuracy at flow speeds of less than 1 m/s, which is a flow regime difficult to measure with enough accuracy for the MAV operation under low-speed wind disturbances.   

\begin{figure}[h!]
      \centering
      \includegraphics[width=0.9\columnwidth]{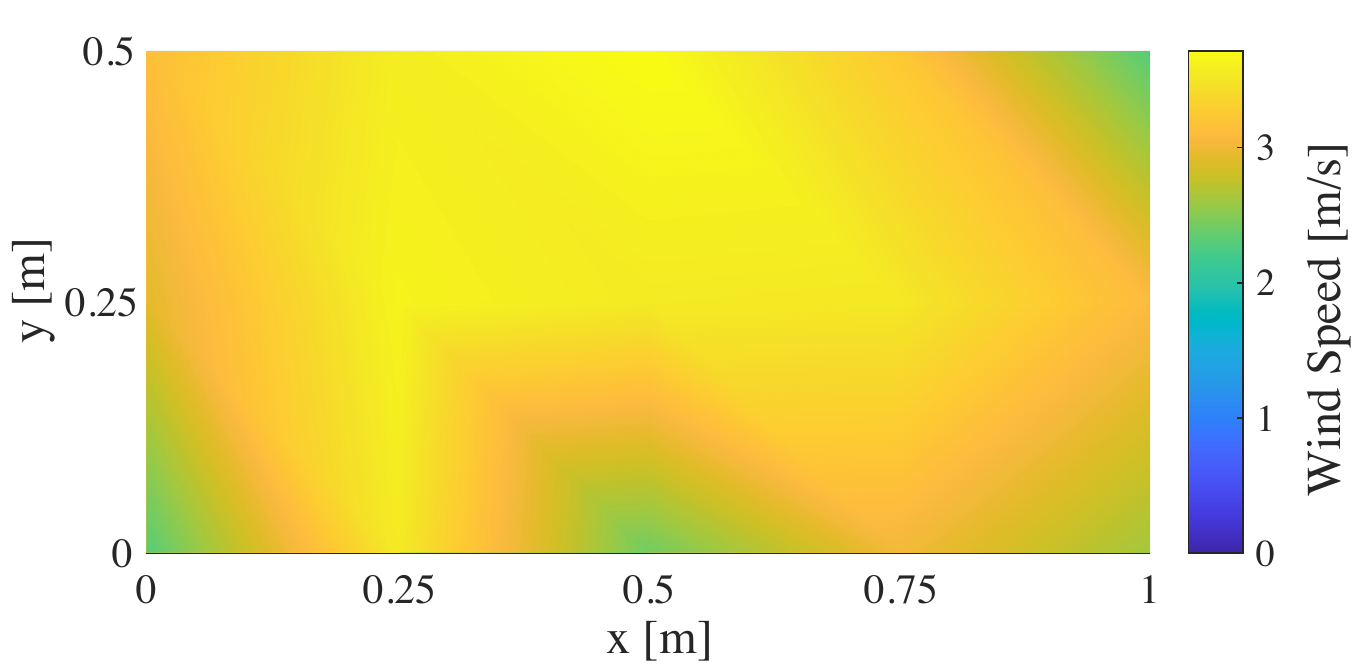}%{RevP_sensor_grid/1_max_RPM_flow_field.eps}
      \caption{Time-averaged wind speed map of the fan system at full RPM at 1 meter from the fan array.}
      \label{fig:flow_map_full_RPM}
      \end{figure}

\begin{figure}[h!]
      \centering
      \includegraphics[width =\columnwidth]{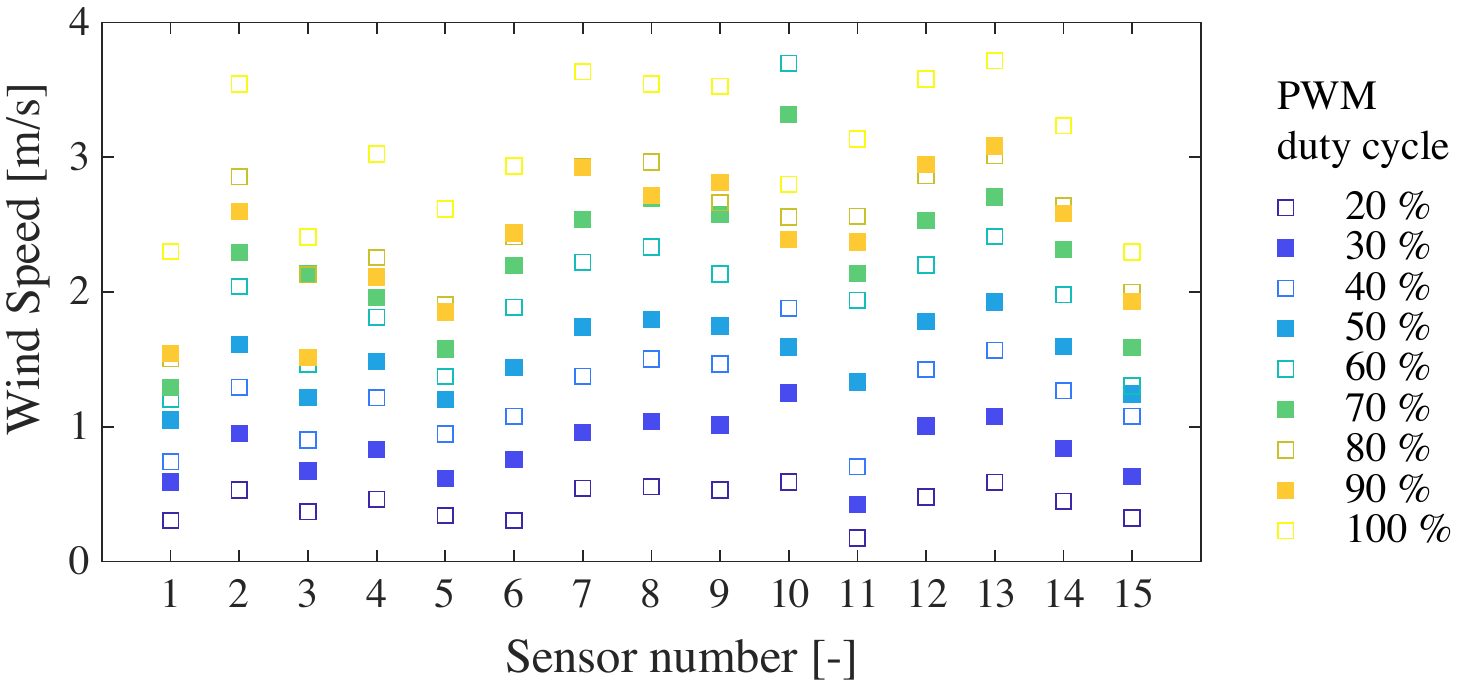}
      \caption{Time-averaged wind speed at various grid points of the fan system with different RPMs, 1 meter from the fan array (The nine different PWM signals here correspond to the RPM settings in Figure \ref{fig:RPM}.)}
      \label{fig:avg_speed_full}
      \end{figure}

The data acquisition of the wind sensing grid is done with two Teensy 4.1 microcontrollers with onboard SD logging, sampling at approximately 3000Hz simultaneously for all 15 sensors for 20 seconds at each fan RPM. Post-processing is done in MATLAB by doing initial sensor offset adjustment and higher-frequency noise removal through low-pass filtering with a cut-off frequency of 50 Hz. Figure \ref{fig:flow_map_full_RPM} shows an example of the interpolated results from the time-averaged wind speeds at the 15 grid points. Overall, reduced flow speeds are observed near the boundary of the flow envelope due to flow expansion and mixing with the ambient, without the use of an encapsulated downstream test section. 
This flow speed fluctuation at the flow volume boundary can also be seen in Figure \ref{fig:avg_speed_full} near sensor locations 1,5,6,10,11,15. By implementing fan tachometer feedback in the wind speed regulation control loop and the possibility of an active wind speed reading with the RevP sensors, the uniformity of the flow field could be improved. The sensing grid's measurement results also provide more accurate time-averaged wind speed mapping of the fan system at various RPMs as shown in Figure \ref{fig:RPM}, given its extra calibration efforts at extremely low flow speeds and higher frequency response compared to the pressure probe used in the ProCap system.

For steady flights, the ideal flight envelope would be in front of the fan modules at the center, where RevP sensors 7,8,9 point towards. This region also has the lowest turbulence intensity around 1.5-5\%, while at the boundary of the flow volume, turbulence intensity could spike up to approximately 22\%. 

\section{Experimental setup}

The wind system was placed in the Cyberzoo, a flight arena (10 m x 10 m x 7 m) of TU Delft equipped with a motion tracking system consisting of 12 OptiTrack Prime 17W cameras. In the center of the fan array, four active markers were added to ensure access to information about changes in wind speed. The markers are programmed to blink with the change of the PWM signal of the fans. The drone enters the flow from the side, then goes to the waypoint located at the 1m distance from the center of the fan array and returns to its starting point. The positioning system, here OptiTrack, sends state data to the drone's ground station. Accurate position control is ensured by the outer-loop guidance algorithm based on proportional−integral−derivative (PID) controllers.

We assumed three sets of experiments, where a drone is flying into the wake of the fan system, gathering data about states, battery voltage and current. Firstly, we focus on steady wind conditions. We set a wind speed and record around 50 seconds of free flight. We repeat the experiment six times for wind speeds ranging from 0.5 m/s to 3.4 m/s. During the second set of experiments, we introduce a wind gust, by periodically modulating wind speed from 1.3 m/s to 3.4 m/s in a loop. The modulation is achieved by repeatedly changing PWM signals from 50$\%$ to 100$\%$ duty cycles at regular intervals. The tested frequencies of wind gust oscillations are 0.5 Hz, 0.25 Hz and 0.125 Hz. 

\subsection{Flapping Wing MAV}
\begin{figure}[h]
      \centering
      \includegraphics[width = 0.7\columnwidth]{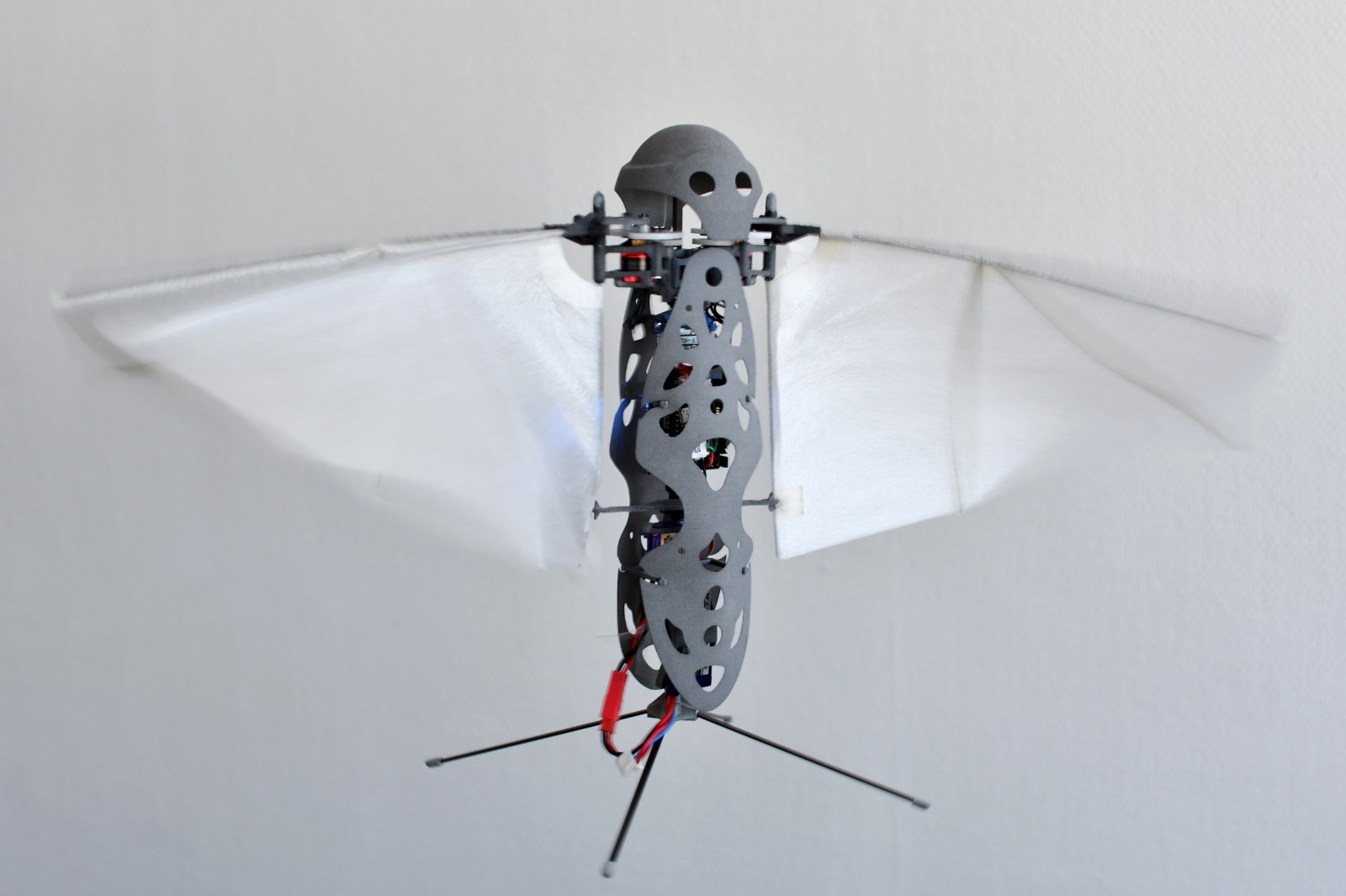}
      \caption{Flapper, a 100-gram bio-inspired flapping-wing MAV from company Flapper Drones. }
      \label{flapper}
   \end{figure}

For the experiments, we have used a flapping wing (Figure \ref{flapper}) delivered by the company Flapper Drones \footnote{https://flapper-drones.com/wp/}, referred to in this paper as "Flapper". This tailless flapping wing MAV is a successor of the DelFly Nimble \cite{c11}. It has a wingspan of 49 cm weights 102g and is capable of carrying 25 g of payload. At hover, the flapping frequency is approximately 12 Hz. The Flapper is equipped with the Crazyflie Bolt autopilot board (Bitcraze AB) and a 6-axis IMU BMI088 - triaxial accelerometer and gyroscope. The data link between the autopilot and the ground station is obtained via Crazyradio PA, a USB radio dongle based on nRF24LU1+ from Nordic Semiconductor. To sustain a 4-6 minutes long flight, depending on the complexity of maneuvers, the robot requires a 300mAh, 7.4V LiPo battery. For the purpose of performing analysis of the onboard power consumption, the ACS711 Hall effect-based linear current sensor was added to the autopilot board.

\subsection{Rotorcraft}
The quadrotor used in this research is the Crazyflie 2.1. This tiny quadrotor built by the company Bitcraze features four brushed motors, an STM32F4 based flight controller \footnote{https://www.bitcraze.io/products/crazyflie-2-1}. The motor-to-motor length is 92mm and the take-off weight with the 3D-printed propellor-guard and added current sensor is 39gr, which is close to the maximum recommended weight of 42gr. The current sensor is the same one as used on the Flapper, the ACS711 linear current sensor. Similar to Flapper, the communication between the Crazyflie 2.1 and the ground station is obtained via the Crazyradio PA. The quadrotor flies with a single cell 250mAh 3.7 LiPo battery, sustaining about 3-4 minutes of flight. 
   
   \begin{figure}[h]
      \centering
      \includegraphics[width = 0.7\columnwidth]{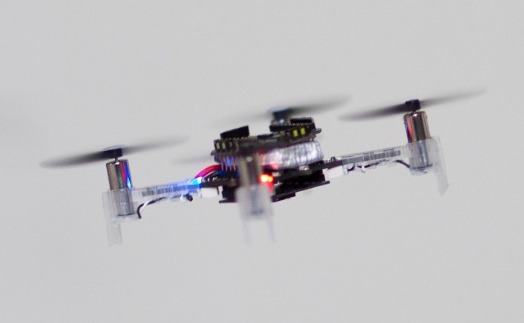}
      \caption{Crazyflie, a 39-gram MAV from company Bitcraze. }
      \label{crazyflie}
   \end{figure}

\section{Results}
\subsection{Various wind speeds}
\begin{figure}[h]
      \centering
      \includegraphics[width =0.9\columnwidth]{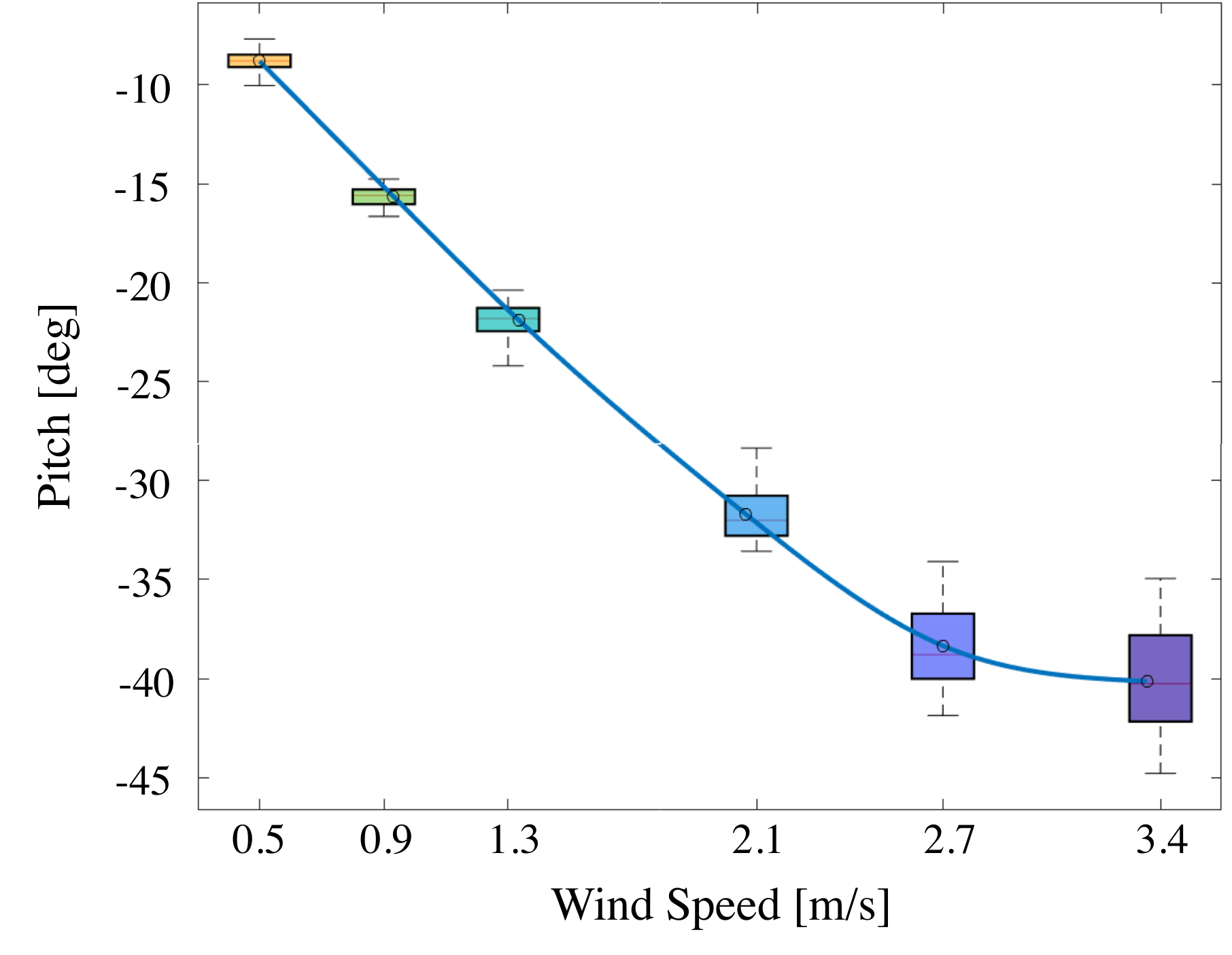}
      \caption{Flapper: Pitch angle at various wind speeds. The average value is marked with a blue circle. Flapper }
      \label{pitchvswindF}
  \end{figure}

\begin{figure}[h]
      \centering
      \includegraphics[width = 0.85\columnwidth ]{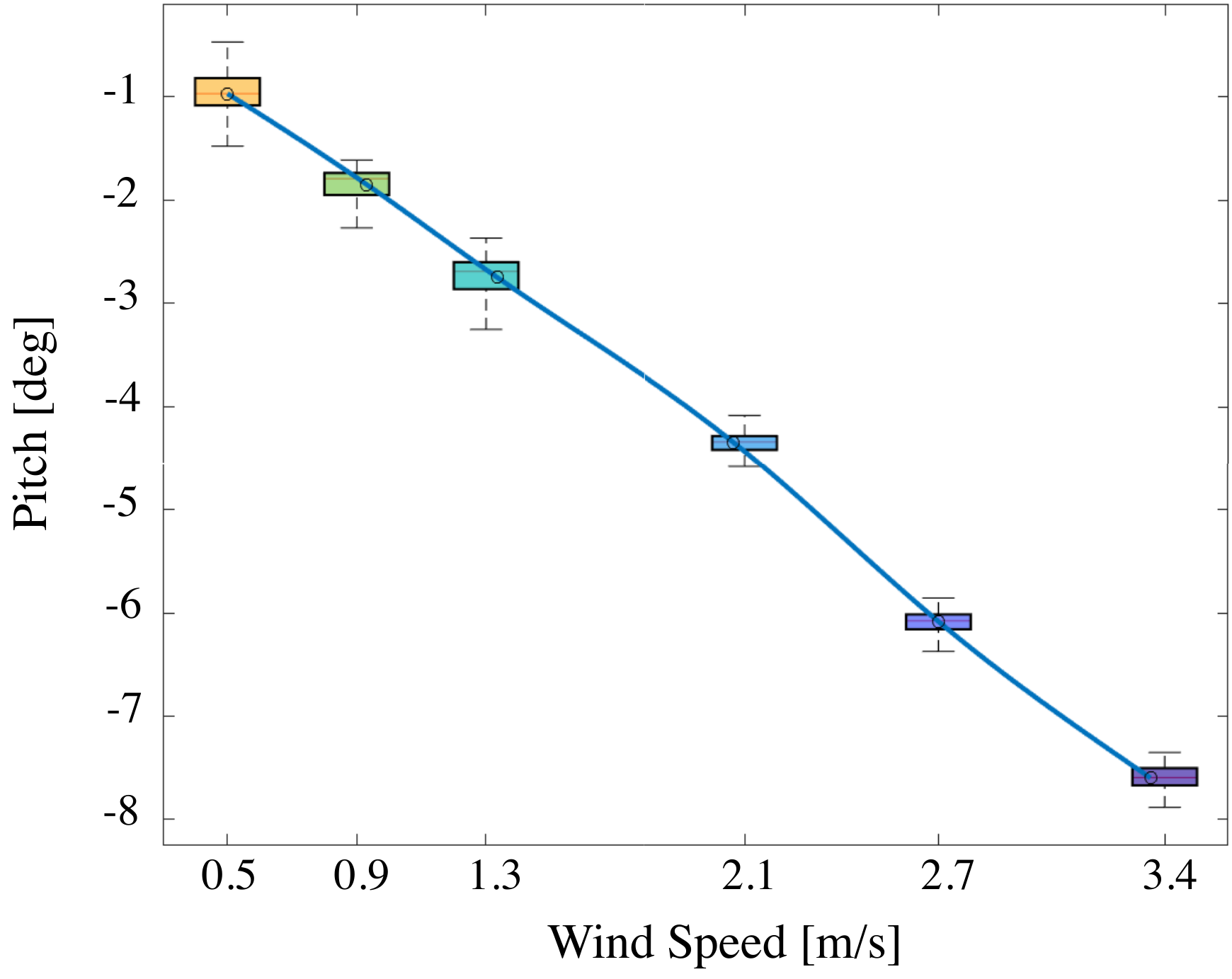}
      \caption{Crazyflie: Pitch angle at various wind speeds. The average value is marked with a blue circle. }
      \label{pitchvswindC}
  \end{figure}

The first set of experiments considers the flight of a MAV at various wind speeds. Figures \ref{pitchvswindF} and  \ref{pitchvswindC} show the relation between the pitch angle and forward speed of the flying MAVs. At the small wind speeds, the MAVs are almost hovering with only a small pitch angle. Once the wind speed increases, the pitch angle becomes more substantial for the Flapper till the point where it reaches a forward flight regime where the wings start to provide lift.

\begin{figure}[h]
      \centering
      \includegraphics[width = 0.9\columnwidth]{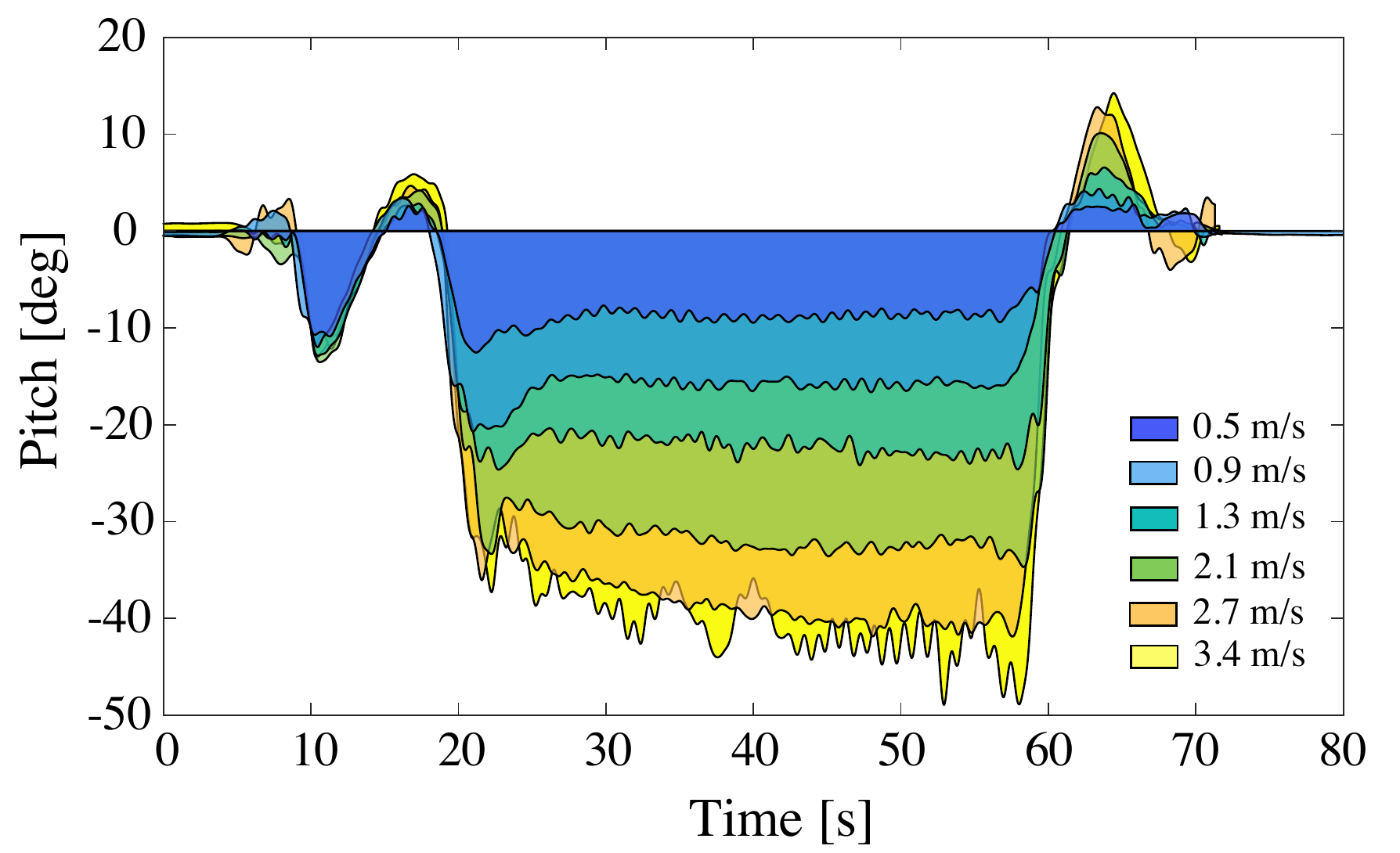}
      \caption{Flapper: Time series of the pitch angle.}
      \label{pitchvstimeF}
   \end{figure}

\begin{figure}[h]
      \centering
      \includegraphics[width = 0.9\columnwidth]{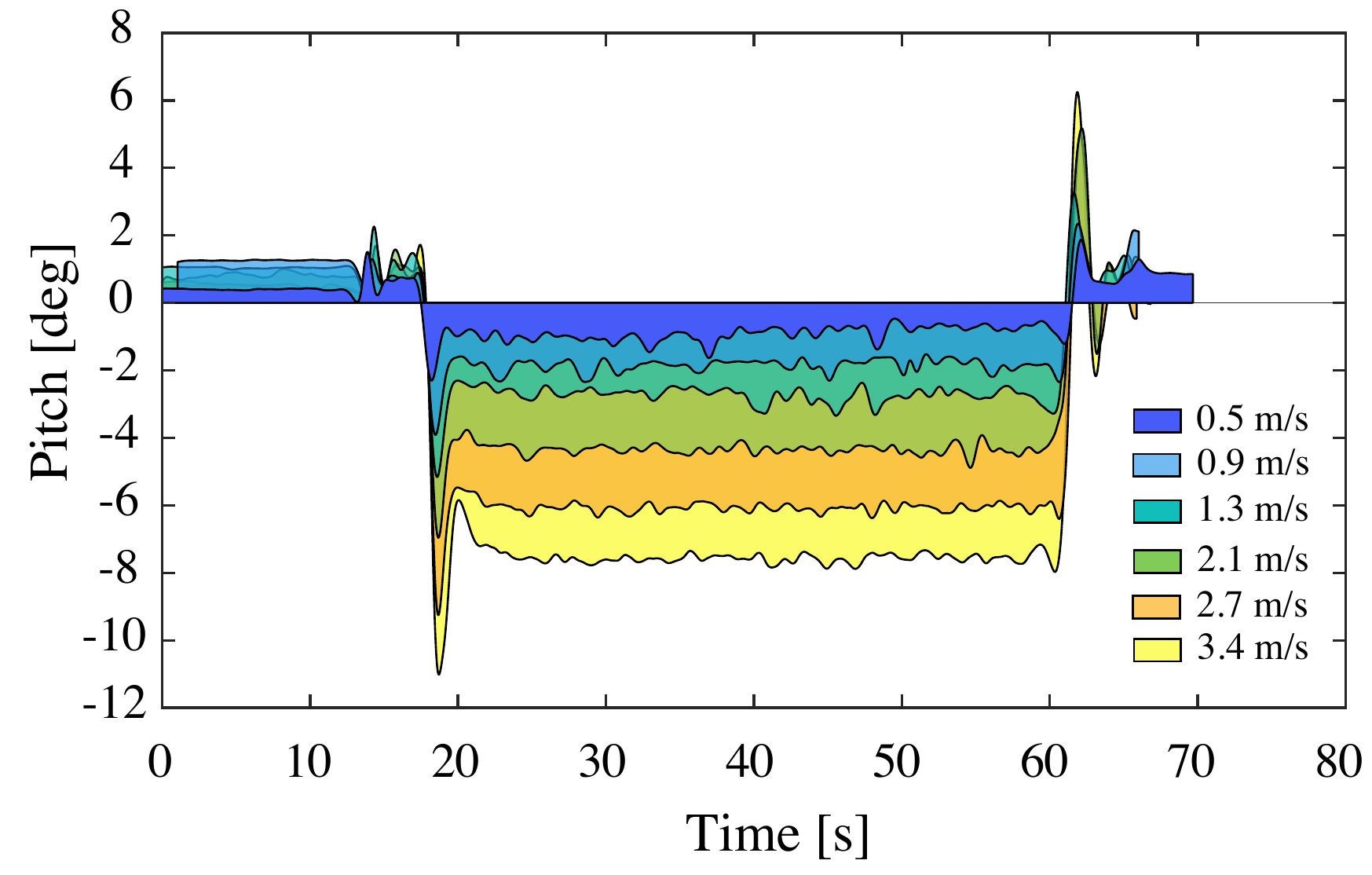}
      \caption{Crazyflie: Time series of the pitch angle.}
      \label{pitchvstimeC}
   \end{figure}

   The magnitude of the pitch angle for the Flapper is much higher compared to the Crazyflie. This is due to the Flapper's much larger wing surface, requiring more thrust to counter the wind. When pitching forward, the wings do provide extra lift, since they function as an airfoil. The time series of pitch angles for both MAVs can be observed in Figures \ref{pitchvstimeF} and \ref{pitchvstimeC}.
   
\begin{figure}[h]
      \centering
      \includegraphics[width = 0.9\columnwidth]{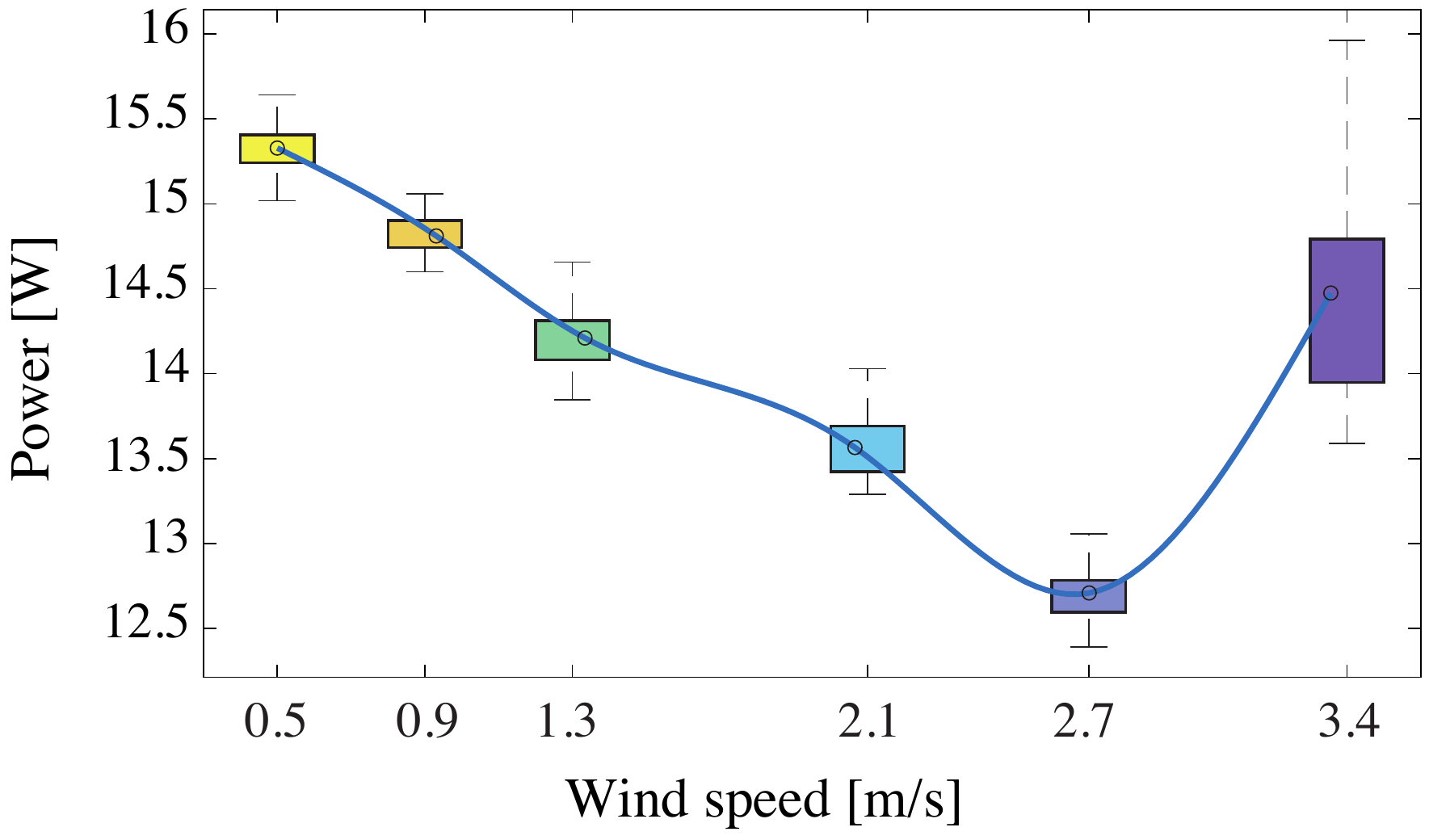}
      \caption{Flapper: Power consumption at various wind speeds. The mean value is marked with a blue circle.}
      \label{powervswindF}
   \end{figure}

\begin{figure}[h]
      \centering
      \includegraphics[width = 0.9\columnwidth]{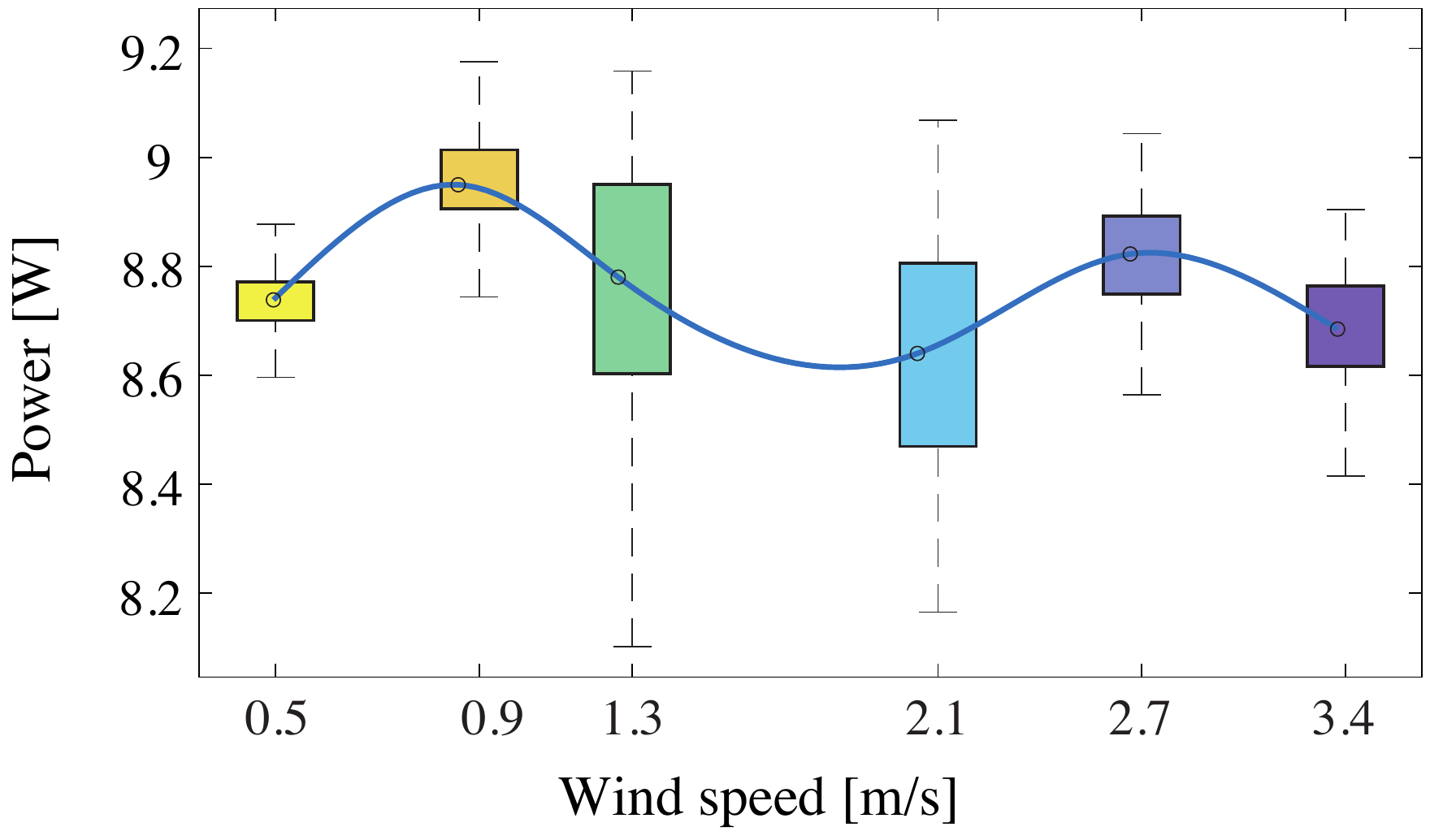}
      \caption{Crazyflie: Power consumption at various wind speeds. The mean value is marked with a blue circle.}
      \label{powervswindC}
   \end{figure}

 Figures \ref{powervswindF} and \ref{powervswindC} show boxplots of the power consumption in function of wind speed, where mean power consumption is approximated using a cubic smoothing spline (blue curve) and the mean values are marked with blue circles. The characteristic bell shape curve of the Flapper drone is reminiscent of the fuel consumption curve of a helicopter \cite{c11}. This indicates that the power consumption is reduced by increasing forward velocity, which is due to the wings providing extra lift. Nevertheless, too high a forward velocity will cause a significant rise in the power demand due to the induced drag. The two tested MAVs are too different for a direct comparison of power consumption (e.g., 102g for the Flapper and 39g for the CrazyFlie). Indeed, in absolute terms, the CrazyFlie uses on average less power, around 8.8W whereas the lowest power consumption of the Flapper is 12.7W, at 2.7m/s wind. However, the trend is indicative of the differences between flapping wing MAVs and quadrotors. Over the varying wind speeds, the CrazyFlie has a much more constant power consumption (Fig.  \ref{powervswindC}). This suggests that more research on the quadrotor's optimal flight speed is needed to demonstrate that dynamic lift reduces power consumption.

  \subsection{Wind gusts}
\begin{figure}[h]
      \centering
      \includegraphics[width = 0.9\columnwidth]{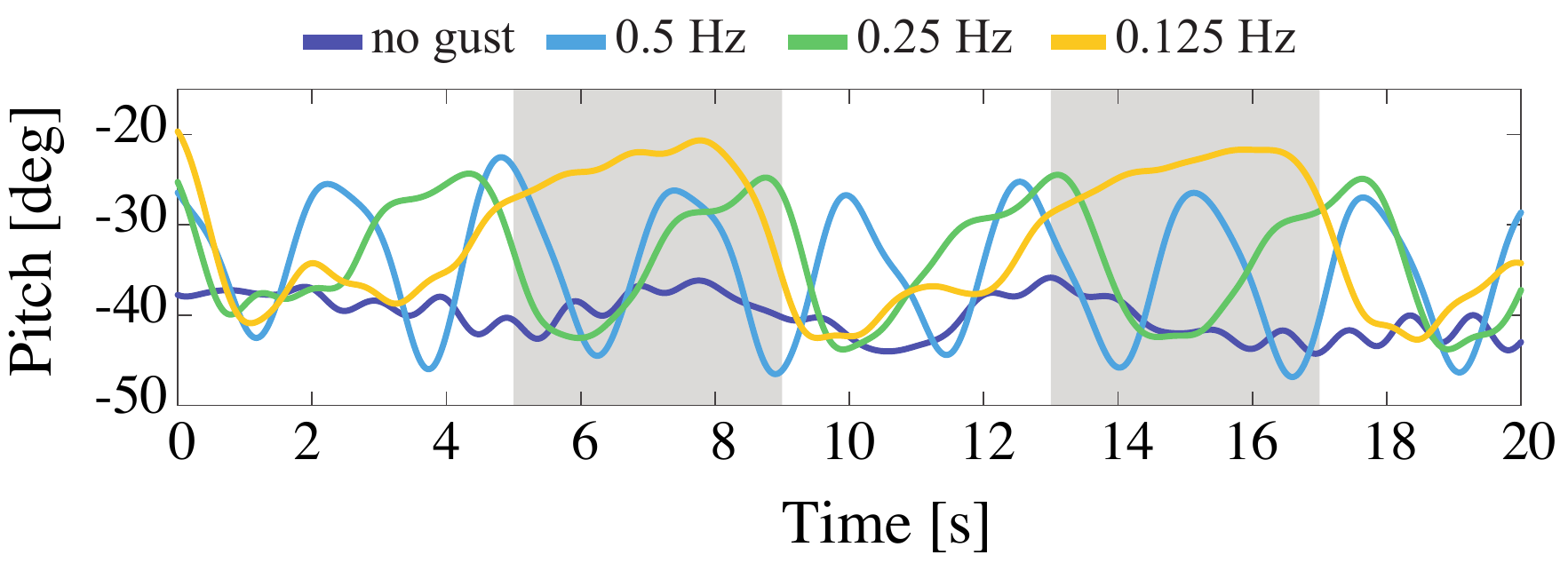}
      \caption{Flapper: Pitch angle at various wind conditions: from steady wind (no gust) to wind gust with various oscillation frequencies ranging from 0.125 Hz to 0.5 Hz.}
      \label{freqF}
   \end{figure}

\begin{figure}[h]
      \centering
      \includegraphics[width = 0.9\columnwidth]{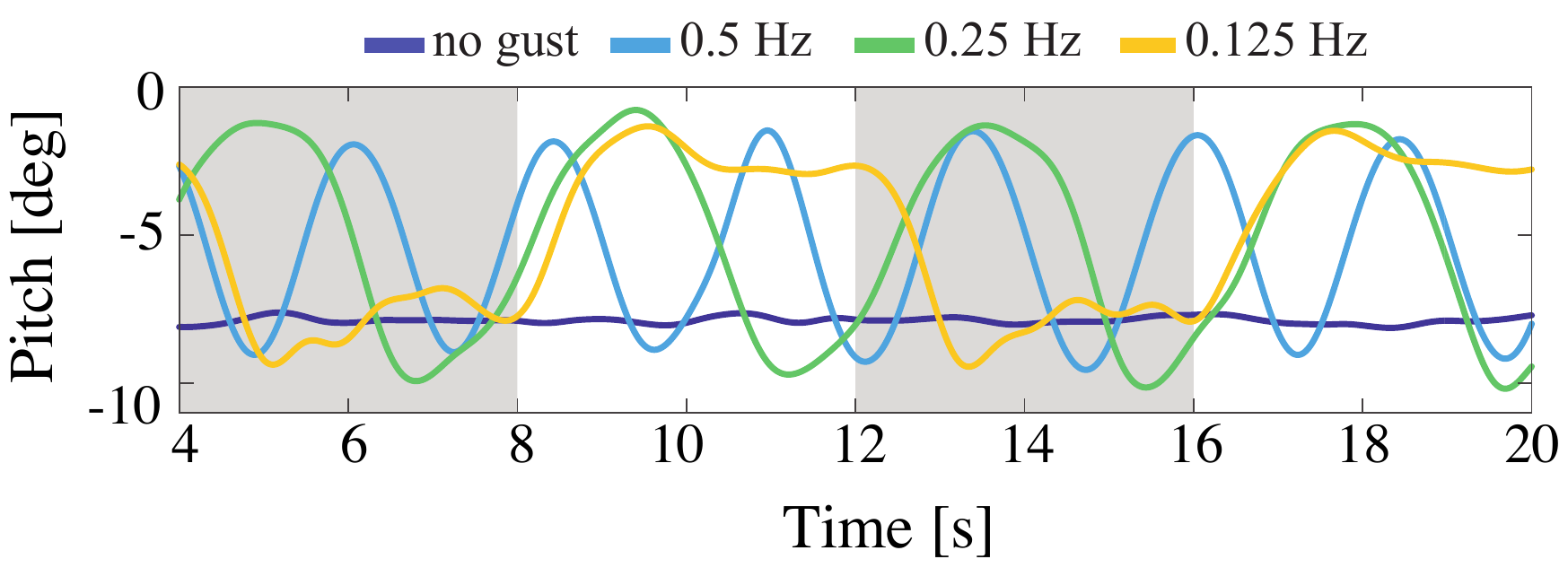}
      \caption{Crazyflie: Pitch angle at various wind conditions: from steady wind (no gust) to wind gust with various oscillation frequencies ranging from 0.125 Hz to 0.5 Hz.}
      \label{freqC}
   \end{figure}

\begin{figure}[h!]
      \centering
      \includegraphics[width = 0.9\columnwidth]{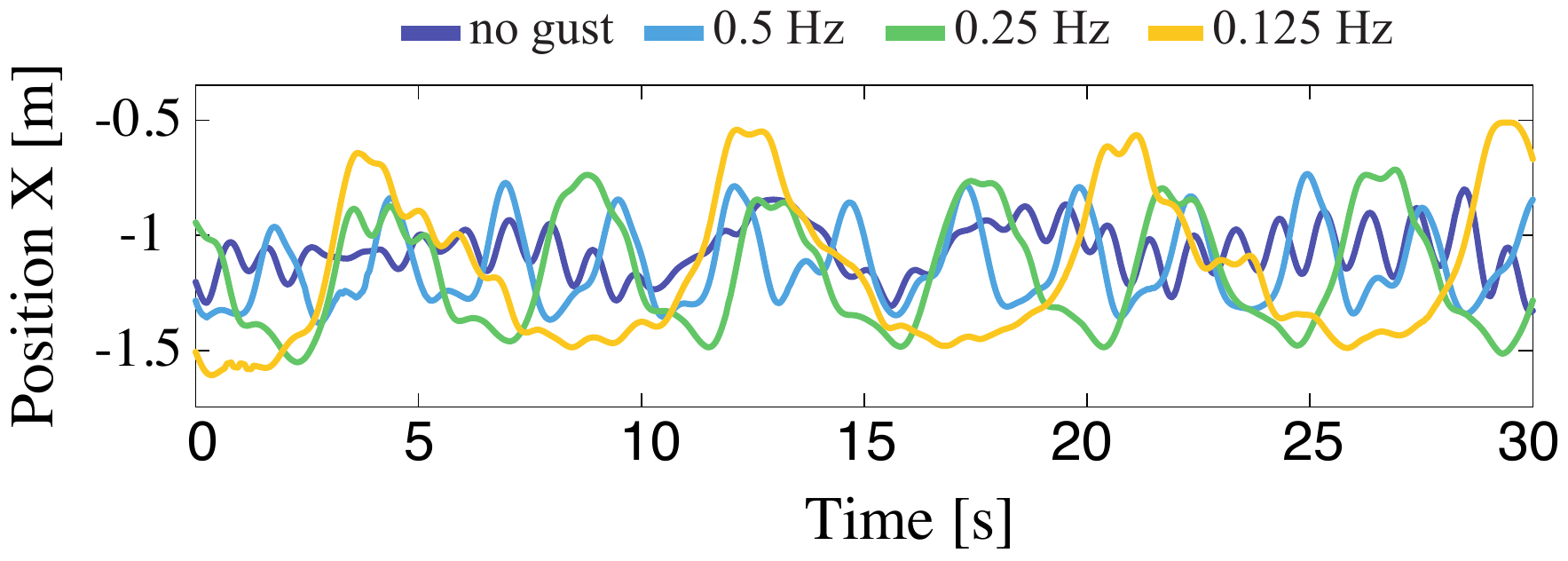}
      \caption{Flapper: Position X at various wind conditions: from steady wind (no gust) to wind gust with various oscillation frequencies ranging from 0.125 Hz to 0.5 Hz.}
      \label{posF}
   \end{figure}

\begin{figure}[h!]
      \centering
      \includegraphics[width = 0.9\columnwidth]{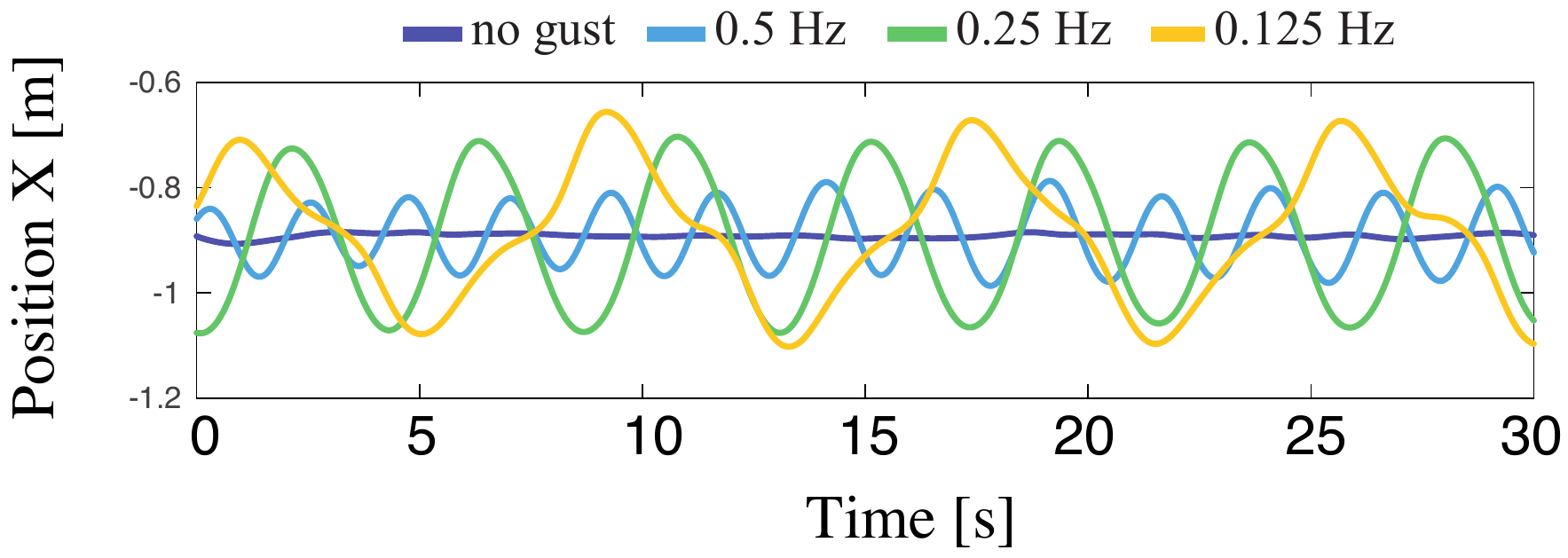}
      \caption{Crazyflie: Position X at various wind conditions: from steady wind (no gust) to wind gust with various oscillation frequencies ranging from 0.125 Hz to 0.5 Hz.}
      \label{posC}
   \end{figure}

  The second set of experiments introduces wind gusts. The wind speed is periodically modulated from 1.3 m/s to 3.4 m/s in a loop. Figures \ref{freqF} and \ref{freqC} show 2 periods of experiment with wind gust oscillation of 0.125 Hz and on top of it 0.25Hz and 0.5Hz. The grey area indicates a change in speed for the 0.125Hz case. The baseline is added - the curve "no gust" which shows flight at steady conditions with a wind speed of 3.4 m/s. The information about the exact moment of the change was retrieved from the active markers mounted on the fan system, which were programmed to blink when a new PWM signal is commanded. The Crazyflie appears to respond faster to changes. The Flapper struggles more to hold the position even during the "no gust" case (See Figures \ref{posF} and \ref{posC}). The Flapper's position control system has a limitation: it commands thrust to go up and pitch to go forward. At the speed 3.4 m/s and at over 40 degrees of pitch, attitude-dependent mapping needs to be added in order to maintain level flight.

%  \subsection{Step response} 
  
 \section{Conclusion}
This work presents the design and characterization of a low-speed, open-source and open-hardware multi-fan wind system. We have performed experiments with two types of MAVs -- a flapping wing and  a quadrotor -- in order to characterize their wind response and power consumption in different constant and varying wind conditions. The experiments showed that flapping wing MAVs have to pitch more in response to oncoming wind, due to the larger wing surface area. However, in higher wind conditions, the wings do provide extra lift, leading to lower power consumption at higher wind speeds. The experiments also showed that the gust responses of both systems should be improved if we are to increasingly employ MAVs in real-world, windy environments. We hope that the proposed open-source, open-hardware fan wall will provide the community with the right tool for developing and testing these improvements.

\end{document}